\documentclass[10pt, a4paper]{article}

\usepackage{lrec-coling2024} 
\usepackage{multirow}
\usepackage{subfigure}
\usepackage{amsmath}
\usepackage{graphicx}
\usepackage{pdfpages}

\title{Difficulty-Focused Contrastive Learning for Knowledge Tracing with a Large Language Model-Based Difficulty Prediction}

\name{Unggi Lee$^{1}$, Sungjun Yoon$^{2}$, Joon Seo Yun$^{2}$, Kyoungsoo Park$^{2}$ \\ {\bf \large YoungHoon Jung$^{2}$, Damji Stratton$^{3}$, Hyeoncheol Kim$^{1}$$^{\dagger}$}}

\address{$^{1}$Korea University, $^{2}$i-Scream Edu, $^{3}$The University of Missouri System \\
         codingchild@korea.ac.kr, {yseong555, jsyun0510, kspark0818, yhjung}@i-screamedu.co.kr, \\ dhsdfn@umsystem.edu, harrykim@korea.ac.kr}

\abstract{
This paper presents novel techniques for enhancing the performance of knowledge tracing (KT) models by focusing on the crucial factor of question and concept difficulty level. Despite the acknowledged significance of difficulty, previous KT research has yet to exploit its potential for model optimization and has struggled to predict difficulty from unseen data. To address these problems, we propose a difficulty-centered contrastive learning method for KT models and a Large Language Model (LLM)-based framework for difficulty prediction. These innovative methods seek to improve the performance of KT models and provide accurate difficulty estimates for unseen data. Our ablation study demonstrates the efficacy of these techniques by demonstrating enhanced KT model performance. Nonetheless, the complex relationship between language and difficulty merits further investigation.
 \\ \newline \Keywords{Knowledge tracing, large language model, contrastive learning} 
}

\begin{document}

\maketitleabstract

\section{Introduction}
Knowledge tracing (KT) is a field of research that aims to predict student learning progress by analyzing their past interactions with question items within an educational context \cite{abdelrahman2023knowledge, corbett1994knowledge}. Difficulty estimation plays a crucial role in understanding dynamic student learning progress \cite{minn2018improving}. Accordingly, developing embeddings adapting item response theory (IRT) models such as the Rasch model has been used to calculate item difficulty \cite{ghosh2020context}. Other studies adapted classical test theory (CTT) to estimate difficulty \cite{lee2022contrastive}.

In addition, contrastive learning has emerged as an effective framework in various research areas, including computer vision, representation learning, as well as KT \cite{2020simclr1, wang2020understanding, lee2022contrastive}. Contrastive learning learns representations by comparing positive and negative samples \cite{wang2020understanding, le2020contrastive}. While there is some previous research on contrastive learning applied to KT, few studies have focused on incorporating the difficulty information to improve model performance. 

Moreover, the textual features of questions in educational contexts contain valuable information about the required skills, question difficulty, and student interaction with questions. According to Abdelrahman, Wang, and Nunes (2023), Deep learning KT models have utilized these textual characteristics to acquire an understanding of question patterns and monitor the levels of knowledge in students. However, the potential role of natural language in KT is not yet fully understood.

The current study aims to address these gaps by proposing a new model, called Difficulty-Focused Contrastive Learning for Knowledge Tracing with a Large Language Model (DCL4KT+ LLM). The model utilizes CTT to calculate concept difficulty and question difficulty and incorporates the contrastive learning framework to enhance the performance of the model. Furthermore, it leverages the textual features of questions to improve the accuracy of knowledge tracing. The architecture of the proposed model consists of embedding layers, encoder blocks based on the MonaCoBERT model, and a contrastive learning framework. In this study, we tested DCL4KT + LLM using the benchmark datasets and compared its performance using AUC and RMSE. Additionally, an ablation study was conducted to examine the effect of difficulty-focused contrastive learning and difficulty prediction using LLM, and the effect of the data augmentation.

\begin{figure*}[t]
\centering
\includegraphics[width=\textwidth]{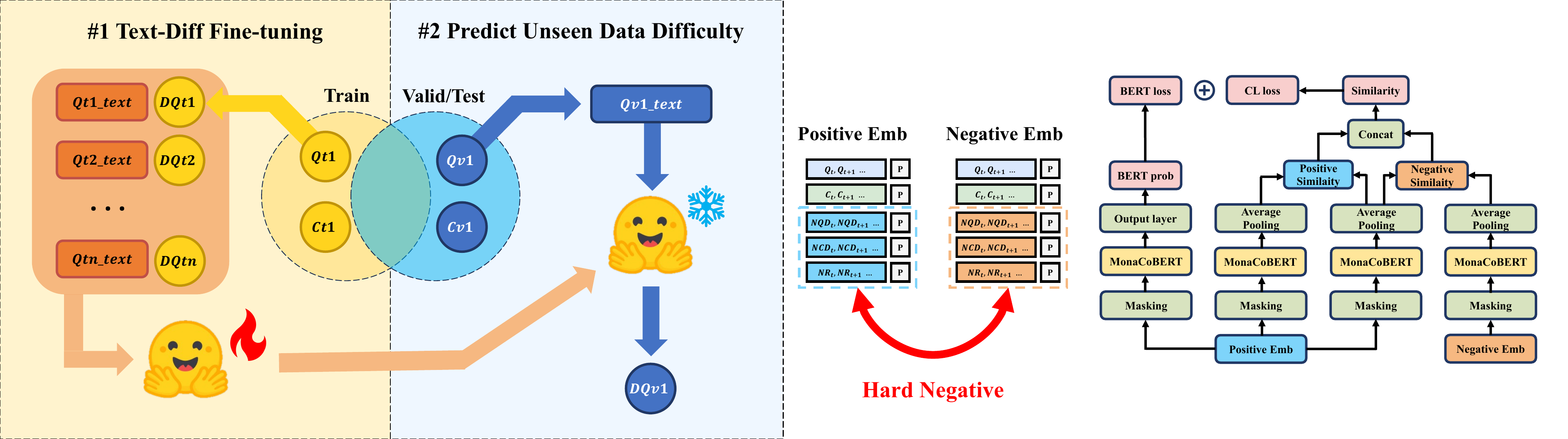}
\caption{Architectures of DCL4KT+LLM. \textit{Left}: LLM-based difficulty prediction framework in KT. \textit{Right}: Whole architecture of DCL4KT+LLM}
\label{whole_arch}
\end{figure*}

\section{Background}

\subsection{Difficulty in Knowledge Tracing}


The difficulty has a significant impact on student learning practices.\cite{minn2018improving}. Previous research in education has explored methods to calculate difficulty in questions or concepts. CTT and IRT are popular methods to calculate difficulty. 

IRT depicts the relationship between an individual's response to an item and their level on the scale's underlying construct. \cite{edelen2007applying}. Attentive knowledge tracing (AKT) \cite{ghosh2020context} used the Rasch embedding strategy to represent difficulties of question and concept, inspired by the Rasch model of item response theory. Other studies have adapted CTT as it is much simpler and thus easier to interpret the results.  \cite{petrillo2015using}. In fact, various KT models use CTT to calculate difficulty; bidirectional encoder representation of knowledge tracing (BEKT) \cite{tianabekt}, monotonic attention-based ConvBERT for knowledge tracing (MonaCoBERT) \cite{lee2022monacobert}, and contrastive learning for knowledge tracing (CL4KT) model \cite{lee2022contrastive}. Meanwhile, the Graph Neural Network (GNN)-based model in KT used a representation of difficulty by using the relationship of questions (concepts) and students' responses \cite{song2022bi, luo2022dagkt}. 

In this research, we used CTT to calculate difficulty and included a concept called 'hard negative difficulty' to empower the performance of our model.

\subsection{Contrastive Learning in Knowledge Tracing}

Recently, contrastive learning-based models have achieved better performance in a lot of research areas, such as computer vision, natural language processing, and recommendation systems. Contrastive learning is a method that learns the representation by comparing the positive samples with the negative samples \cite{wang2020understanding, le2020contrastive}. Momentum Contrast for Unsupervised Visual Representation Learning (MOCO) \cite{moco2022} proposed a dynamic dictionary using a queue and moving-averaged encoder, which improved the performance in unsupervised visual representation tasks. SimCLR \cite{2020simclr1, 2020simclr2} achieved better performance by using data augmentation for contrastive learning. Yet, there have been few KT studies that have utilized the  contrastive learning framework. CL4KT \cite{lee2022contrastive} employed contrastive learning in KT by using reversed answer data as negative samples and suggested several data augmentation techniques. In addition, there have been attempts to combine contrastive learning and GNN \cite{song2022bi, wu2023self, dai2022contrastive}. Nonetheless, there are only a few contrastive learning-based KT studies that have explored the role of difficulty in enhancing model performance.

\subsection{Knowledge Tracing with Natural Language Dataset}

The text of a question can contain a great wealth of information, such as the skills required by the question, the difficulty of the question, and the relationships between questions. Several deep learning KT models have leveraged the textual features of question texts to learn question representations and track students' knowledge states \cite{abdelrahman2023knowledge}.

Relation-aware self-attention for knowledge tracing (RKT) and hierarchical graph knowledge tracing (HGKT) also extract features from the textual information of questions to learn question representations in their models. Exercise-enhanced recurrent neural network (EERNN) and exercise-aware knowledge tracing (EKT) extended from EERNN proposed a framework that considers both exercising records and the texts of exercises for predicting student performance \cite{su2018exercise, liu2019ekt}. Adaptable knowledge tracing (AdaptKT), which transfers knowledge from the source domain to the target one, and In Exercise Hierarchical Feature Enhanced Knowledge Tracing utilize Bert has been proposed
\cite{cheng2022adaptkt, tong2020exercise}. QuesNet is also an unsupervised learning method that leverages a large corpus of unlabelled questions
\cite{yin2019quesnet}.

Yet, previous research has not considered much about the latent representation of the textual features in the questions and concepts. Text-aware KT models are motivated by leveraging the textual features of questions and concepts to enhance performance in tackling KT tasks.

\section{Methodology}

\subsection{Problem Statement}

By analyzing the sequence of interaction data collected from a learning management system (LMS) or intelligent tutoring system (ITS), KT attempts to predict the likelihood of a student answering accurately. Student interactions can be represented as $x_1,..., x_t$. Each interaction in KT consists of three components: the query id, the related educational concept, and the student's response. $x_t = (c_t, q_t, r_t)$ describes the $t$-th interaction. $c_t$ represents the educational concept associated with the $t$-th inquiry in this equation. The $q_t$ variable represents the question's identifier. $r_t$ represents the student's response to the $t$-th query, where $r_t$ in $0, 1$, where $0$ denotes an incorrect response and $1$ denotes a correct response. Difficulties can be divided into two; concept difficulties $cd_t$ and question difficulties $qd_t$. The difficulty was set to an integer value ranging from 0 to 100. Based on classical test theory (CTT), The formula for calculating difficulty is the number of students who got the question (concept) correct divided by the total number of questions (concepts).

\subsection{Proposed Model Architecture}

\subsubsection{Embedding Layers}

DCL4KT uses a positive embedding layer block and a negative embedding layer. The positive embedding layer $E_{positive}$ consists of element-wise embedding layers; questions $E_q$, concepts $E_c$, question difficulties $E_{qd}$, concept difficulties $E_{cd}$ and students' response $E_r$. In addition, position embedding $E_{p}$ also contained the positive embedding block. The formulation of the positive embedding layer is below.

\begin{equation}
E_{positive} = E_q + E_c + E_{qd} + E_{cd} + E_r + E_{p}
\end{equation}

The negative embedding layer $E_{negative}$ consisted of element-wise embedding layers; questions $E_q$, concepts $E_c$, hard negative question difficulties $E_{nqd}$, hard negative concept difficulties $E_{ncd}$ and hard negative students' response $E_{nr}$. The formulation of the negative embedding layer is

\begin{equation}
E_{negative} = E_q + E_c + E_{nqd} + E_{ncd} + E_{nr} + E_{p}
\end{equation}
The position embedding $E_{p}$ also contained the negative embedding block. The details of the negative embedding are shown in section 3.3.2.

\subsubsection{Encoder Architecture}

In this research, we used MonaCoBERT \cite{lee2022monacobert} as an encoder block. MonaCoBERT is a transformer-based model which changes the attention module by combining with span-based dynamic convolution (SDC) and monotonic attention (MA), which model can represent students' response sequence locally and globally while representing the students' forgetness. DCL4KT uses four MoncaCoBERT encoder modules where each module used four transformer layers $Tr$. Three encoder modules are used for the contrastive learning framework. One encoder calculates binary cross entropy (BCE) loss, and three encoders calculate contrastive learning loss.

\subsection{Contrastive Learning Framework}

\subsubsection{Loss Function}

The loss function of DCL4KT is calculated by summing the BCE loss $\mathcal{L}_{bce}$ and contrastive loss $\mathcal{L}_{cl}$ \cite{lee2022contrastive}. The ratio between BCE loss and contrastive loss is controlled by the hyper-parameter $\lambda_{c}$ which ranges from $[0, 1]$. The whole loss function is formulated as

\begin{equation}
\mathcal{L} = (1 - \lambda_{c}) \times \mathcal{L}_{bce} + \lambda_{c} \times \mathcal{L}_{cl},
\end{equation}

BCE loss $\mathcal{L}_{bce}$ is a binary cross entropy loss between prediction $\hat{r}_t$ and real students' response $r_t$, defined as 

\begin{equation}
\mathcal{L}_{bce} = \sum_t-\left(r_t \log \hat{r}_t + \left(1-r_t\right) \log \left(1-\hat{r}_t\right)\right)
\end{equation}

Contrastive loss $\mathcal{L}_{cl}$ is a concatenate of concept similarity $sim_{c}$ and question similarity $sim_{q}$,

\begin{equation}
\mathcal{L}_{cl} = concat(sim_{c}, sim_{q})
\end{equation}


When positive concept pair $c_{t1}^+, c_{t2}^+$ is passed through the encoder layer $tr$, the result is $cz_{t1}^+, cz_{t2}^+ = tr(c_{t1}^+, c_{t2}^+)$. When positive and negative concept pair $c_{t1}^+, c_{t2}^-$ is passed through the encoder layer $tr$, the result is $cz_{t1}^+, cz_{t2}^- = tr(c_{t1}^+, c_{t2}^-)$. Thus, the concept similarity is defined as

\begin{equation}
sim_{c}=-\log \frac{e^{\operatorname{sim}\left(cz_{t1}^+, cz_{t2}^+\right)}}{e^{\operatorname{sim}\left(cz_{t1}^+, cz_{t2}^+\right)}+\sum{\operatorname{sim}\left(cz_{t1}^+, cz_{t2}^-\right)}}
\end{equation}

When positive question pair $q_{t1}^+, q_{t2}^+$ is passed through the encoder layer $tr$, the result is $qz_{t1}^+, qz_{t2}^+ = tr(q_{t1}^+, q_{t2}^+)$. When positive and negative concept pair $q_{t1}^+, q_{t2}^-$ is passed through the encoder layer $tr$, the result is $qz_{t1}^+, qz_{t2}^- = tr(q_{t1}^+, q_{t2}^-)$, such that

\begin{equation}
sim_{q}=-\log \frac{e^{\operatorname{sim}\left(qz_{t1}^+, qz_{t2}^+\right)}}{e^{\operatorname{sim}\left(qz_{t1}^+, qz_{t2}^+\right)}+\sum{\operatorname{sim}\left(qz_{t1}^+, qz_{t2}^-\right)}}
\end{equation}

\subsubsection{Embedding with Hard Negative}

The novel implementation of the contrastive learning framework in CL4KT \cite{lee2022contrastive} included negative embedding for student responses. Our research expands hard negative embedding to question and concept difficulty. The positive and negative embeddings are depicted on the right Figure \ref{whole_arch}.

The positive embedding is a composition of question components $E_q$, concepts $E_c$, question difficulty $E_{qd}$, conceptual difficulty $E_{cd}$, and student responses $E_r$. Negative embedding, on the other hand, integrates element-wise combinations of question components $E_q$, concepts $E_c$, hard negative of question $E_{nqd}$, concept difficulty $E_{ncd}$, and hard negative student responses $E_{nr}$.

To provide additional clarity, the hard negatives are derived in a particular manner. For example, if a student's answer is correct, i.e. $1$, the corresponding hard negative becomes $0$. In contrast, if a student's response is incorrect i.e. $0$, the hard negative is marked as $1$. Similarly, concerning difficulty, if the difficulty of a query or a concept is rated at $0.75$, then its hard negative equivalent would be $0.25$. If the difficulty rating is $0.25$, the corresponding negative value is $0.75$. This can be summarized as

\begin{equation}
E_{nqd} = 1 - E_{qd}, E_{ncd} = 1 - E_{cd}, E_{nr} = 1 - E_r,
\end{equation}
where $E_{qd}, E_{cd}, E_r, E_{nqd}, E_{ncd}, E_{nr}$ is $[0, 1]$.

\subsection{LLM-based Difficulty Prediction Framework}

In KT, when we calculate difficulty from questions and concepts, it is not possible to calculate the difficulty of the dataset, which is contained in the validation and test dataset but not in the training dataset, due to data splitting. Previous KT Model with difficulty used human-selected hyper-parameters \cite{lee2022monacobert} or used representations of Question and Concept \cite{ghosh2020context, lee2022contrastive}. However, this approach is not stable in realistic educational Intelligent Tutoring Systems (ITS) or online learning platforms, which consistently add new questions or concepts to the e-learning system. This need led to the creation of a new approach to predict the difficulty of unseen questions or concepts which are not contained in the training dataset.

We present an LLM-based difficulty prediction framework to calculate difficulty, which is contained in the validation and test datasets but not in the training dataset, using the text of questions and concepts. The left side of Figure \ref{whole_arch} shows the LLM-based difficulty prediction framework.

We define the notation to formulate the LLM-based difficulty prediction framework.

\begin{itemize}
    \item $D$: a dataset of students' responses in knowledge tracing.
    \item $D_{train}$, $D_{valid}$, $D_{test}$: subsets of $D$ representing the training, validation, and test sets, respectively.
    \item $d(q, c)$: difficulty score for a question $q$ and a concept $c$.
    \item $B_{pr}$: a pre-trained BERT model.
    \item $B_{ft}$: a fine-tuned BERT model.
\end{itemize}

First, the dataset $D$ consists

\begin{equation}
D = {(q_i, c_i, r_i)}_{i=1}^{N},
\end{equation}

where $q_i$ is the $i$-th question, $c_i$ is the ith concept, and $r_i$ is the $i$-th response. And we split the dataset into training, validation, and testing sets as

\begin{equation}
    D_{train}, D_{valid}, D_{test} = Split(D, ratio)
\end{equation}

where $Split$ is a function that divides the dataset based on a specific ratio.

Then, we calculate the difficulty $CalDiff$ scores from the training set, not the validation and test set as

\begin{equation}
    d(q_i, c_i) = CalDiff(D_{train})
\end{equation}

Using the $d(q_i, c_i)$, we execute fine-tune, $FT$, the pre-trained BERT model, which is trained by text corpus. In this research, we used pre-trained KoBERT \footnote{https://huggingface.co/beomi/kobert} downloaded from Huggingface \footnote{https://huggingface.co/}, because our dataset contained Korean text, not English. The formulation is below.

\begin{equation}
B_{ft} = FT(B_{pr}, d(q_i, c_i)), 
\end{equation}

where $FT$ is a function that updates the model parameters using the training dataset and the calculated difficulties.

The fine-tuned BERT model is used to predict the difficulties of questions and concepts in the validation/test sets,

\begin{equation}
    \hat{d}(q_j, c_j) = B_{ft}(q_j, c_j)
\end{equation}

\begin{equation}
    \forall (q_j, c_j) \in D_{valid} \cup D_{test},  \quad \forall (q_j, c_j) \notin D_{train}
\end{equation}

where $\hat{d}(q, c)$ is the predicted difficulty of questions and concepts in the validation/test sets which are not contained in the training sets.

\subsubsection{Data Augmentation}

Referencing the previous research in KT and NLP, we developed and applied eleven data augmentation strategies for DCL4KT. To control the probability of application, we set the probability hyper-parameter to each augmentation strategy. If the hyper-parameter $\gamma_{crop}$ is 0.2, then the probability of application crop is 20\%. All of the data augmentation strategies are applied to the training session, not the validating or testing session.

\begin{itemize}

    \item \textbf{Token cutoff, span cutoff \cite{shen2020simple}}: The token cutoff is a simple augmentation technique that removes random data portions from an input sentence to produce limited perspectives. As a variant of the cutoff procedure, span cutoff removes a continuous segment of text.
    \item \textbf{Concept and question mask \cite{lee2022contrastive}}: This method masks the concept or question randomly. The probability is the same as the original BERT. Note that the MonaCoBERT encoder already uses the students' correctness mask.
    \item \textbf{Crop \cite{lee2022contrastive}}: A method which crops the parts of the sequence.
    \item \textbf{Summarize}: Maintains the order of the sequence and extracts some elements in the sequence.
    \item \textbf{Reverse}: Reverses the order of elements in the sequence.
    \item \textbf{permute \cite{lee2022contrastive, yang2019xlnet}}: Permutes the order of elements in the sequence randomly.
    \item \textbf{Segment permute}: Makes segments, then permutes those segments.
    \item \textbf{Replace higher and lower difficulty \cite{lee2022contrastive}}: Replaces questions or concepts up to the difficulty.
    \item \textbf{Concatenate sequence}: Concatenates two sequences to make new sequences.

\end{itemize}

\begin{table*}[hbt!]
    \renewcommand{\arraystretch}{1.5}
    \centering
    \begin{tabular}{ccccccccccc}
        \hline
        Dataset & Metrics & DKT & DKVMN & AKT & CL4KT & MCB-C & DCL4KT & DCL4KT-A \\
        \hline
        \multirow{2}{*}{ASSISTments09} & AUC & 0.7285 & 0.7271 & 0.7449 & 0.7600 & 0.8059 & \underline{0.8111} & \textbf{0.8153} \\
                            & RMSE & 0.4328 & 0.4348 & 0.4413 & 0.4337 & \underline{0.4063} & 0.4068 & \textbf{0.4034}  \\
        \hline
        \multirow{2}{*}{Algebra05} & AUC & 0.8088 & 0.8146 & 0.7673 & 0.7871 & 0.8201 & \underline{0.8288} & \textbf{0.8295}  \\
                            & RMSE & 0.3703 & 0.3687 & 0.3918 & 0.3824 & \textbf{0.3584} & 0.3657 & \underline{0.3644}  \\
        \hline
        \multirow{2}{*}{Algebra06} & AUC & 0.7939 & 0.7961 & 0.7505 & 0.7789 & 0.8064 & \underline{0.8258} & \textbf{0.8278}  \\
                            & RMSE & 0.3666 & 0.3661 & 0.3986 & 0.3863 & 0.3672 & \underline{0.3522} & \textbf{0.3504}  \\
        \hline
        \multirow{2}{*}{EdNet} & AUC & 0.6609 & 0.6602 & 0.6687 & 0.6651 & 0.7336 & \underline{0.7392} & \textbf{0.7403}  \\
                            & RMSE & 0.4598 & 0.4597 & 0.4783 & 0.4750 & 0.4516 & \underline{0.4505} & \textbf{0.4500}   \\
        \hline
        \multirow{2}{*}{Homerun20} & AUC & 0.7619 & 0.7543 & 0.5903 & 0.6014 & 0.7659 & \underline{0.7766} & \textbf{0.7808}  \\
                            & RMSE & 0.4092 & 0.4212 & 0.4745 & 0.4631 & 0.4880 & \underline{0.4042} & \textbf{0.4014}   \\
        \hline
    \end{tabular}
    \caption{Overall performance of KT models based on four benchmark datasets and one custom dataset. The best performance is denoted in bold, and the second is underlined. DCL4KT-A indicates DCL4KT that used augmentation strategies. We can see that DCL4KT-A achieved the best results, and DCL4KT was second for most of the benchmark datasets.}
    \label{tb:performance}
\end{table*}

\subsection{Experiment Setting}

\subsubsection{Datasets}

\begin{itemize}

    \item \textbf{ASSISTment09}: The ASSISTment datasets were collected from the ASSISTment intelligent tutoring system (ITS), predominantly from middle schools in the U.S., with participants randomly assigned \cite{heffernan2014assistments}. We used ASSISTments09 and ignored ASSISTments15 which does not contain question information\footnote{retrieved from https://sites.google.com/site/assistmentsdata/home}.
    
    \item \textbf{Algebra05, 06}: The algebra datasets, provided by the KDD Cup 2010 Educational Data Mining Challenge, were collected from Cognitive Tutor. This ITS, developed by Carnegie Learning, focuses on middle school students \cite{ritter2007cognitive}\footnote{retrieved from https://pslcdatashop.web.cmu.edu/KDDCup}.
    
    \item \textbf{EdNet}: Ednet dataset is provided by an edtech company in South Korea named Santa with a primary focus on the English test TOEIC presented by ETS. It consists of a total of 131,441,538 interactions, accumulated from 784,309 students since 2017, primarily targeting adult learners who need to certify their English competency \cite{choi2020ednet}\footnote{retrieved from https://github.com/riiid/ednet}. We extracted 5,000 interaction data from the original dataset.

    \item \textbf{Homerun20}: The open-source data in KT were not contained full of text about questions and concepts. Because of that, we used the homerun20 dataset, which is not published. This dataset is owned by i-Scream Edu which is an edTech company in South Korea. We used 351,425 responses from 201 elementary school students who used i-Scream Homerun\footnote{https://www.home-learn.co.kr/main/Index.do} math education service in 2020. We erased all personal identity information (PII) before using this dataset. We used de-identified user id, questions and concepts id, the text of questions and concepts, and timestamp.
\end{itemize}

\subsubsection{Evaluation Metrics and Validation}

We employed AUC and RMSE as performance metrics. In addition, we utilized a five-fold cross-validation in our evaluation.

\subsubsection{Baseline Models}

We compared DCL4KT and DCL4KT-A to the baseline models, such as DKT \cite{piech2015dkt}, DKVMN \cite{zhang2017dynamic}, SAKT \cite{pandey2019self}, and the latest models, such as AKT \cite{ghosh2020context}, CL4KT \cite{lee2022contrastive} and MonaCoBERT \cite{lee2022monacobert}.

\subsubsection{Large Language Models}

While developing DCL4KT, we used KoBERT. In ablation studies, we used three LLMs; KoBERT, KoElectra and KoBigbird. These models are trained by Korean corpus. We fine-tune these models using Huggingface transformers to predict the difficulties of unseen data.

\subsubsection{Hyperparameters for Experiments}
To compare each model, we used the same parameters for the model training.
 
\begin{itemize}
\item \textbf{Batch size}: The batch size was 512. Owing to a limitation of resources, we also used a gradient accumulation.
\item \textbf{Early stop}: The early stop parameter was 10. For example, if the validation score was not successively increased during the ten iterations, the training session was stopped.
\item \textbf{Training, validation, test ratio}: The training ratio was 80\% of the entire dataset, and the test ratio was 20\%. The valid ratio was 10\% of the training ratio.
\item \textbf{Learning rate and optimizer}: The learning rate was 0.001, and Adam was used as the optimizer.
\item \textbf{embedding size}: The embedding size was 512.

\item \textbf{Contrastive learning ratio}: We used a contrastive learning ratio as 0.1.

\item \textbf{Augmentation setting}: For augmentation, we set the probability option to control the application of augmentation. Mask-prob is 0.2, crop-prob is 0.1, summarize-prob is 0.2, reverse-prob is 0.1, permute-prob is 0.1, segment-permute-prob is 0.1, replace-higher-diff-prob is 0.1, replace-lower-diff-prob is 0.1, concat-seq-prob 0.1. Also, we used cut-off, not span-cut-off, and the cut-off-prob is 0.03.

\item \textbf{Others}: We used eight attention heads. The max sequence length was 100, and the encoder number was 4. Models used for comparison, such as AKT\footnote{https://github.com/arghosh/AKT} and CL4KT\footnote{https://github.com/UpstageAI/cl4kt}, used the default settings.
\end{itemize}

\begin{figure*}[t]
    \centering
    \begin{subfigure}
        \centering
        \includegraphics[width=0.3\linewidth]{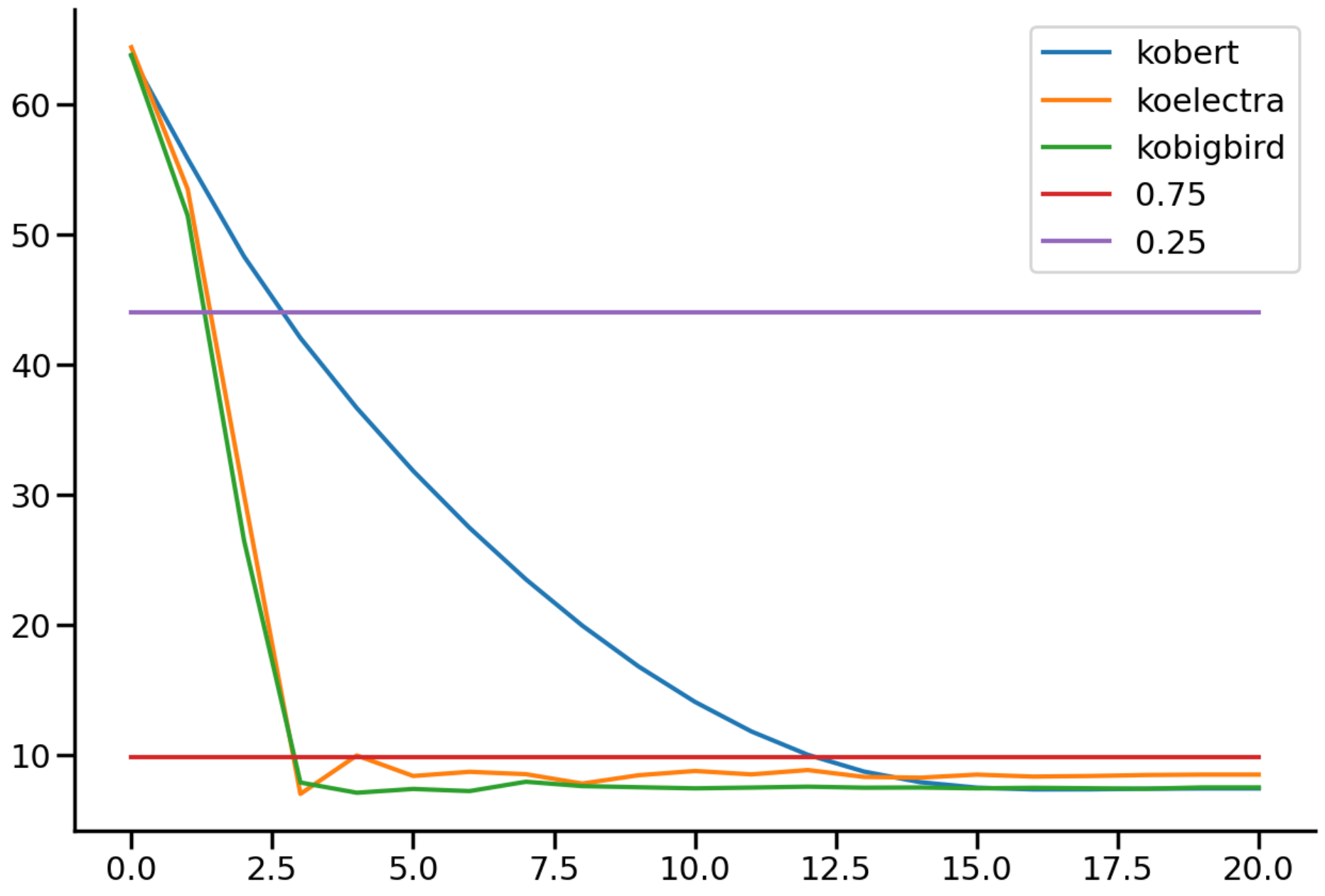}
    \end{subfigure}
    \hfill
    \begin{subfigure}
        \centering
        \includegraphics[width=0.3\linewidth]{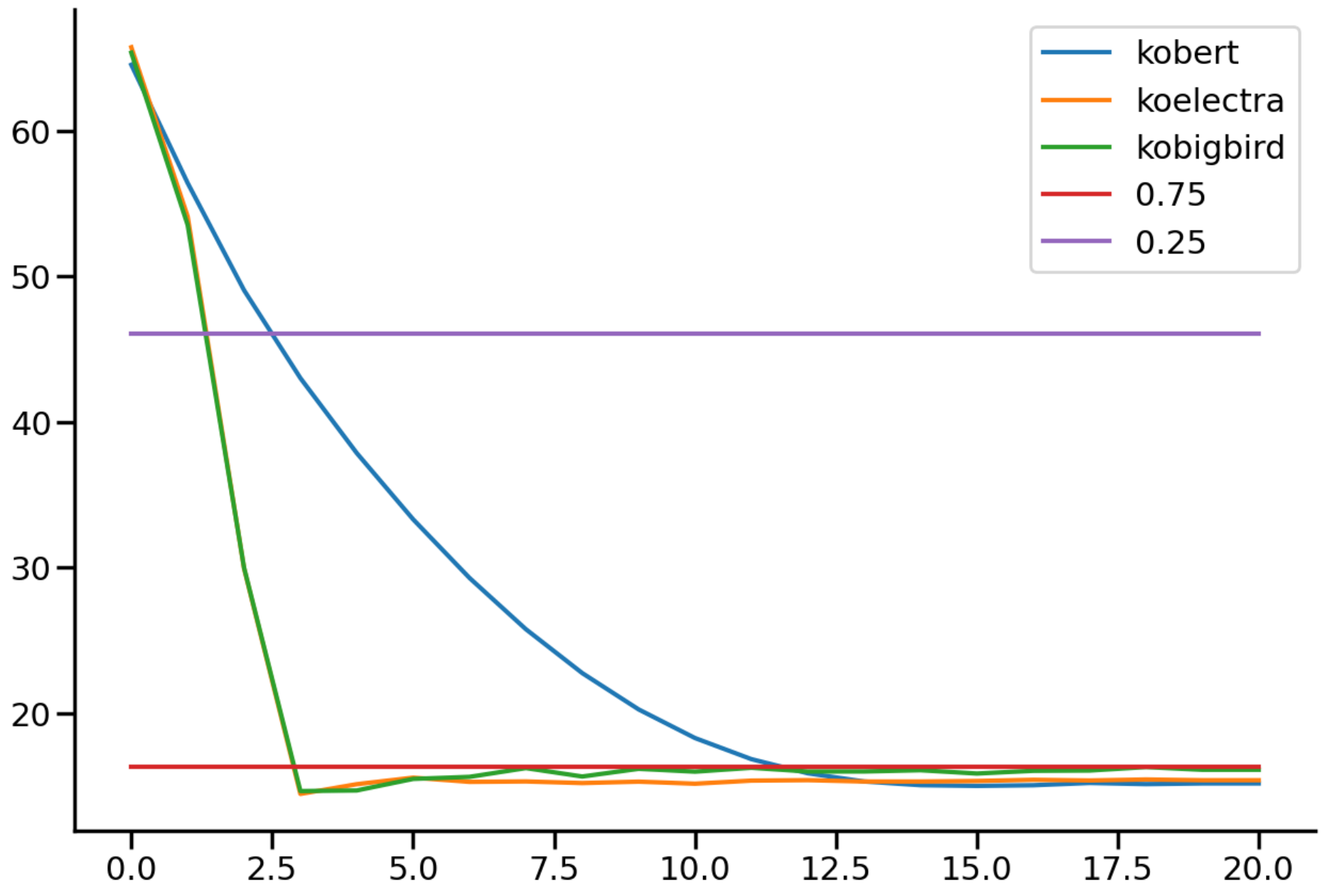}
    \end{subfigure}
    \hfill
    \begin{subfigure}
        \centering
        \includegraphics[width=0.3\linewidth]{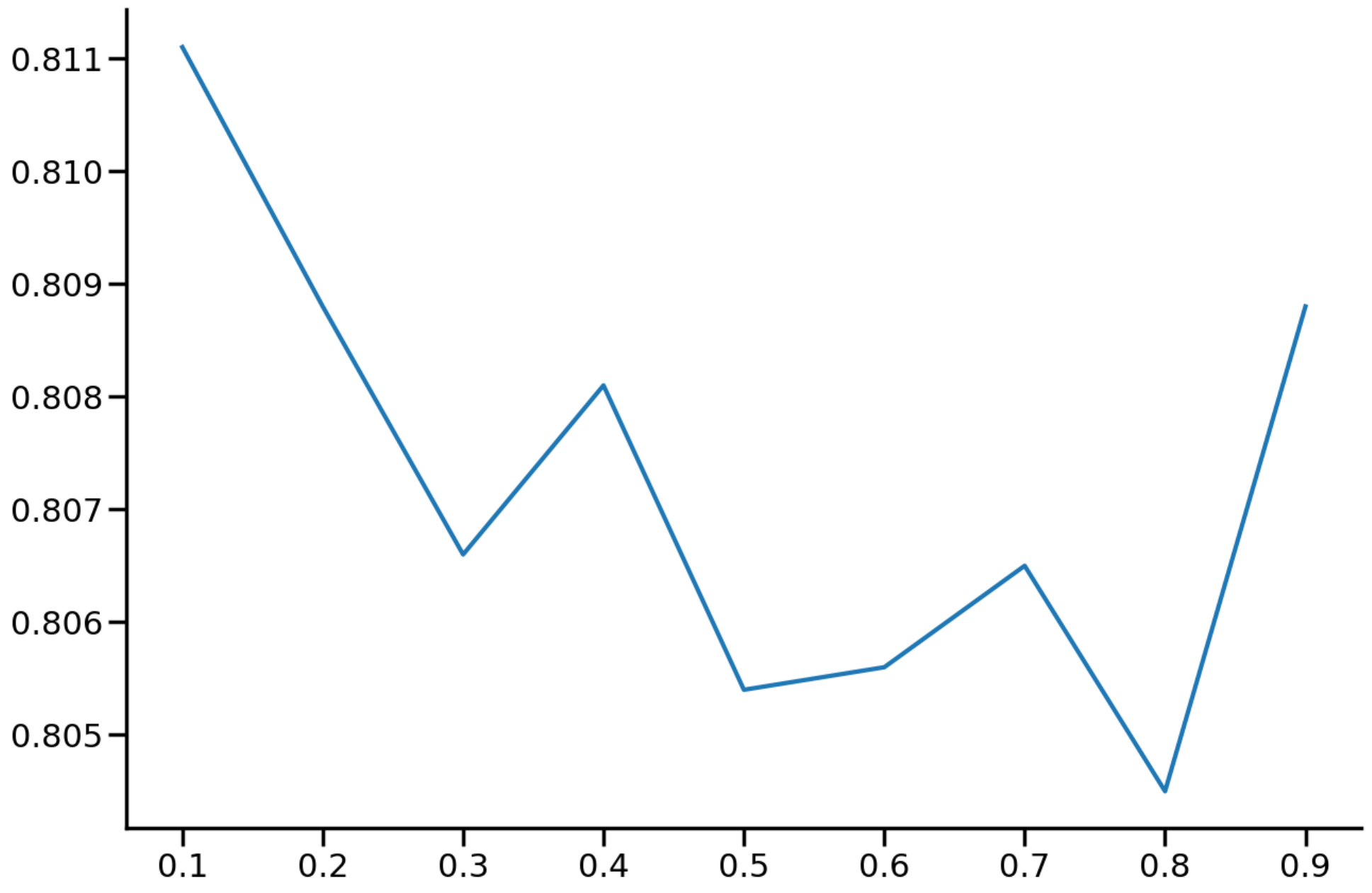}
    \end{subfigure}
    \caption{
        \textit{Left}: Concept difficulty prediction. \textit{Center}: Question difficulty prediction between LLMs. The \textit{x-axis} is training step and \textit{y-axis} means RMSE score. The RMSE score of LLMs are lower than hyper-parameter 0.75. That means LLMs can predict difficulty by using text data of questions and concepts. \textit{Right}: Relationship between contrastive learning ratio (\textit{x-axis}) and model's AUC score (\textit{y-axis}).
    } 
    \label{tb:ablation}
\end{figure*}

\section{Result and Discussion}

\subsection{Overall Performance}

We estimated the overall performance of KT models based on four benchmark datasets and one custom dataset. Table \ref{tb:performance} show the performance of each model. Except for the algebra05 (RMSE), DCL4KT-A achieved the highest performance in all of the benchmark datasets. DCL4KT-A is the version where the augmentation strategies are applied to DCL4KT (The hyper-parameter setting of DCL4KT-A can see the \textbf{Ablation Studies - Effect of Data Augmentation}). The performance of DCL4KT also followed DCL4KT-A.

\subsection{Ablation Studies}

\subsubsection{Effect of Difficulty-focused Contrastive Learning}

To investigate the effect of difficulty-focused contrastive learning, we compared the performance of two cases; 1) non-difficulty-focused contrastive learning (\textit{Non-Diff-CL}), 2) difficulty-focused contrastive learning (\textit{Diff-CL}). For the \textit{Non-Diff-CL} case, the difficulty level of 0.75 is applied to all unseen data in both positive and negative embeddings. Meanwhile, for the \textit{Diff-CL} case, the difficulty level is at 0.75 for positive embedding and 0.25 for negative embedding. As a result, \textit{Diff-CL} achieved higher performance on all of the benchmark datasets. The result is summarized in Table \ref{td:diff-cl}.

\subsubsection{Difficulty Prediction using LLM} \label{subsection-difficulty-prediction}

Using the RMSE metric, we compared two hyper-parameters and three LLMs to determine whether LLMs are capable of predicting difficulty. The two hyper-parameters we selected were 0.75 and 0.25, respectively, representing the average difficulty. Using 0.75 as the hyper-parameter in DCL4KT, the model performed optimally based on our provided data. In contrast, the performance was at its lowest when the hyper-parameter was set to 0.25. Consequently, if the LLM prediction score is near the hyper-parameter value of 0.75, the LLM can effectively replace the use of hyper-parameters and heuristics. KoBERT, KoElectra, and KoBigbird were each trained on a Korean corpus to align with our Korean text dataset, and subsequently evaluated for use in our experiment.

As a result, the left plot in Figure \ref{tb:ablation} shows the concept difficulty prediction. The hyperparameter 0.75 scores 9.8455 and 0.25 scores 44.0410. Meanwhile, KoBERT scores 7.4140, Koelectra scores 8.4801, and KoBigbird scores 7.4923. The center plot in Figure \ref{tb:ablation} shows the question difficulty prediction. The hyperparameter 0.75 scores 16.3373 and 0.25 scores 44.0410. The LLMs also score better than the hyperparameter 0.75. KoBERT scores 15.1846, Koelectra scores 15.4034, and KoBigbird scores 16.1202. These indicate that LLMs can predict difficulty and our proposed LLM-based difficulty prediction framework works effectively on real data. Moreover, we can assume the corpus of problems or concepts has information related to the difficulties, and difficulty can be represented by the corpus.

\begin{table}[t]
\renewcommand{\arraystretch}{1.5}
\centering
\begin{tabular}{cccc}
    \hline
    Dataset & Metric & \textit{Non-Diff-CL} & \textit{Diff-CL} \\
    \hline
    \multirow{2}{*}{ASSISTments09} & AUC & 0.8080 & 0.8111 \\
    & RMSE & 0.4070 & 0.4068 \\
    \hline
    \multirow{2}{*}{Algebra05} & AUC & 0.8223 & 0.8288 \\
    & RMSE & 0.3721 & 0.3657 \\
    \hline
    \multirow{2}{*}{Algebra06} & AUC & 0.8254 & 0.8258 \\
    & RMSE & 0.3525 & 0.3522 \\
    \hline
    \multirow{2}{*}{EdNet} & AUC  & 0.7357 & 0.7392 \\
    & RMSE & 0.4598 & 0.4505 \\
    \hline
\end{tabular}
\caption{Comparing performance of \textit{Non-Diff-CL} and \textit{Diff-CL}. \textit{Non-Diff-CL} is applied difficuty as 0.75 to all of the unseen data. Meanwhile, \textit{Non-Diff-CL} is applied up to positive embedding (0.75) and negative embedding (0.25). The performance of \textit{Diff-CL} is better than the \textit{Non-Diff-CL}.}
\label{td:diff-cl}
\end{table}

\subsubsection{Contrastive Learning Loss Ratio}

We experiment how the contrastive learning framework affects the performance of the model (AUC). We used DCL4KT, a model to which augmentation strategies have not been applied, and the ASSISTments09 dataset for comparison. The right plot in Figure \ref{tb:ablation} \textit{right} shows the relationship between contrastive learning loss ratio (\textit{x-axis}) and the model's performance (\textit{y-axis}). When the contrastive learning loss ratio is 0.1, the performance is best (0.8111). Meanwhile the contrastive learning loss ratio is 0.8, the performance is worst (0.8045).

\subsubsection{Effect of Data Augmentation}

\begin{figure}[]
    \centering
    \includegraphics[width=\linewidth]{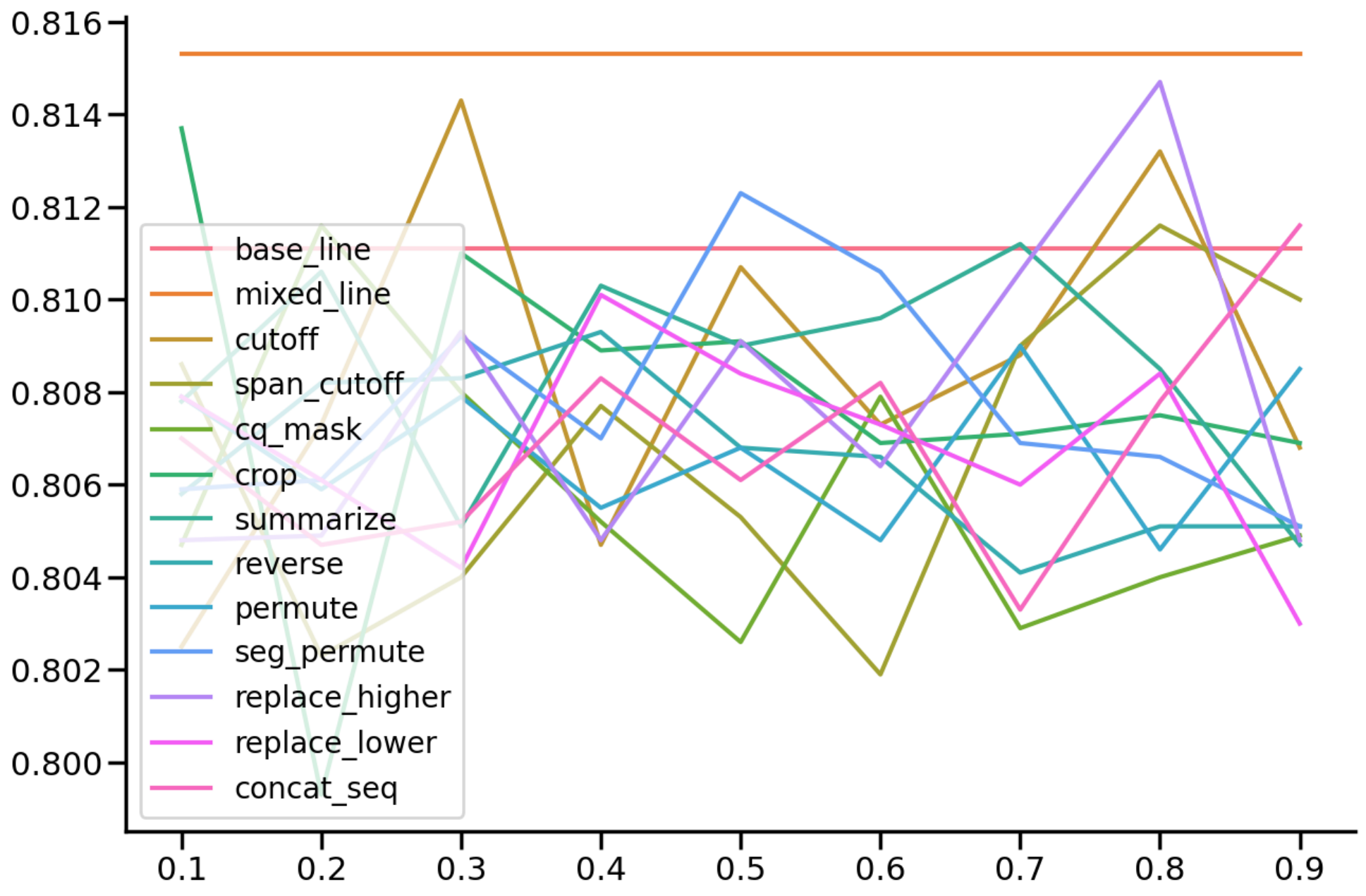}
    \caption{
         Comparing data augmentation strategies. The \textit{x-axis} is data augment probabilities and \textit{y-axis} means AUC score. The baseline is non-augmented DCL4KT.
    } 
    \label{tb:augment}
\end{figure}

We estimated AUC score of eleven augmentation strategies, shown in (Figure \ref{tb:augment}), on the ASSISTments09 dataset. Each augmentation strategy is applied independently. The baseline is non-augmented DCL4KT (0.8111). Some of the augmentation strategies are higher than baseline; \textit{cutoff}, \textit{span cutoff}, \textit{replace higher difficulty}. However, when we estimate performance of mixed augmentation, the probabilities are higher (0.8153) than performance of each augmentation independently. Our hyperparameter settings are as follows: \textit{mask-prob} is 0.2, \textit{crop-prob} is 0.2, \textit{summarize-prob} is 0.2, \textit{reverse-prob} is 0.2, \textit{permute-prob} is 0.3, \textit{segment-permute-prob} is 0.2, \textit{replace-higher-diff-prob} is 0.3, \textit{replace-lower-diff-prob} is 0.2, \textit{concat-seq-prob} is 0.2. We used \textit{cutoff} instead of \textit{span-cutoff}, \textit{cutoff-prob} is 0.03. Note that this setting is not optimized, and therefore, there is room to increase the performance of the model.

\subsection{Relationship between language and difficulty}

\begin{figure}[]
    \centering
    \includegraphics[width=\linewidth]{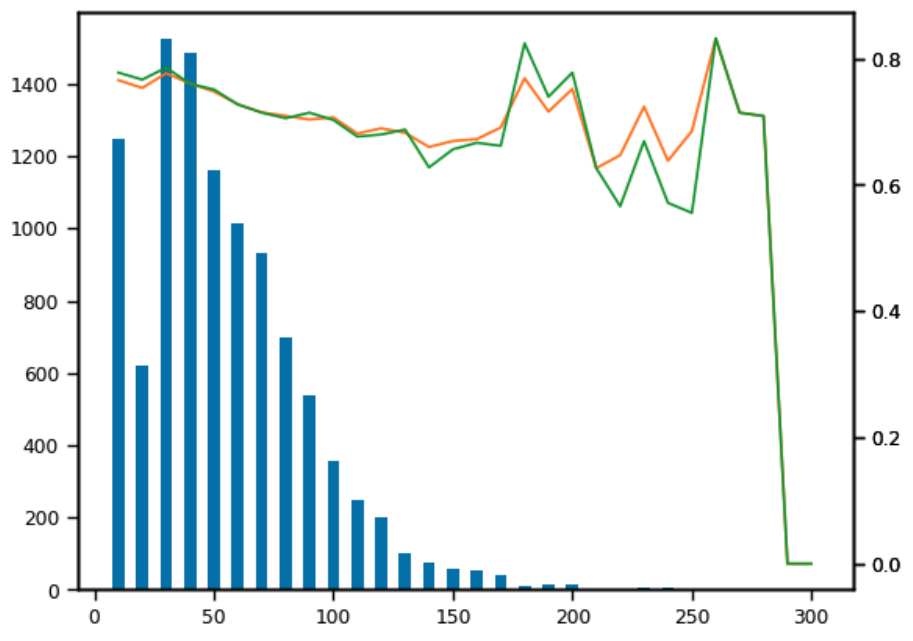}
    \caption{
         Relationship between character length and difficulty. \textit{x-axis} is the character count of questions. \textit{y-axis left} and \textit{blue histogram} mean the number of character length in the dataset. \textit{y-axis right} mean difficulty. \textit{orange line} is mean of correctness, \textit{green line} is median of correctness. When the character count is less than 120, we can see that the students' correctness decreases as the character length increases.
    } 
    \label{fig:relationship}
\end{figure}

In the section titled \textbf{Estimating Difficulty with LLM}, we demonstrated the predictive potential of LLM based on text data. This indicates that the language of text data contains inherent information about its difficulty. To delve deeper into the relationship between language and difficulty, we examined the relationship between variables derived from text data and difficulty.

Figure \ref{fig:relationship} displays the correlation between character count and difficulty. The \textit{x-axis} represents the character length of queries, whereas the \textit{y-axis} left and \textit{blue histogram} represent the character count within the dataset. The phrase \textit{y-axis right} refers to adversity. The \textit{orange line} represents the mean level of correctness, while the \textit{green line} represents the median level of correctness. Due to the relatively small number of queries exceeding 120 characters, we only considered instances where the character count was less than 120. The graph depicts a decline in students' accuracy as character length increases, which holds for both the mean and median correctness.

The character count extracted from text data can be regarded as one of the hidden variables influencing the text's difficulty level. Nonetheless, a more comprehensive examination with additional data from various disciplines must confirm the above findings.

\section{Conclusion}

The significance of difficulty level on student learning habits and the efficacy of the KT model is noteworthy. However, previous KT research has yet to exploit difficulty to improve performance fully and has also struggled to calculate difficulty in unseen data. In response to these obstacles, we have developed a difficulty-centered contrastive learning technique for KT models and a Large Language Model (LLM)-based difficulty prediction framework. These novel techniques can optimize the performance of the KT model and estimate the difficulty level of unknown data. Our ablation investigation confirmed the efficacy of these new techniques for improving the KT model. Nonetheless, the relationship between language and difficulty requires additional study. In subsequent research, we intend to identify the linguistic characteristics that possibly indicate difficulty level.


\section{References}\label{sec:reference}



\end{document}